\documentclass[11pt]{article}
\usepackage{eamt23}
\usepackage{times}
\usepackage{url}
\usepackage{latexsym}
\usepackage[small,bf]{caption} 
\usepackage{graphicx}
\usepackage{longtable}
\usepackage{xcolor}
\usepackage{booktabs}

\title{Reducing Hallucinations in Neural Machine Translation with Feature Attribution}

\author{Joël Tang\\
  Imperial College London\\
  {\tt joel.tang20@imperial.ac.uk} \And
  Marina Fomicheva\\
  University of Sheffield\\
  {\tt m.fomicheva@sheffield.ac.uk}  \AND
  Lucia Specia\\
  Imperial College London\\
  {\tt l.specia@imperial.ac.uk}}
\date{}

\begin{document}
\maketitle
\begin{abstract}
Neural conditional language generation models achieve the state-of-the-art in Neural Machine Translation (NMT) but are highly dependent on the quality of parallel training dataset. When trained on low-quality datasets, these models are prone to various error types, including {\it hallucinations}, i.e.  outputs that are fluent, but unrelated to the source sentences. These errors are particularly dangerous, because on the surface the translation can be perceived as a correct output, especially if the reader does not understand the source language. We present a case study focusing on model understanding and regularisation to reduce hallucinations in NMT. We first use feature attribution methods to study the behaviour of an NMT model that produces hallucinations. We then leverage these methods to propose a novel loss function that substantially helps reduce hallucinations and does not require retraining the model from scratch.
\end{abstract}

\section{Introduction}

Despite impressive progress in Machine Translation (MT) with neural approaches \cite{bahdanau2014neural,vaswani2017attention}, translation adequacy, i.e. the preservation of the source meaning in the translated sentence, still constitutes an issue in current models \cite{castilho2017neural,sharou-specia-2022-taxonomy}. In particular, such systems may ``hallucinate'' content, i.e. generate outputs that read fluently  in the target language, but are completely unrelated to the source sentences \cite{muller-etal-2020-domain,raunak-etal-2021-curious}. These errors are particularly dangerous as they can go unnoticed by the end user of MT if they do not speak the source language.

While there has been substantial work on improving NMT quality in general, very few previous studies have looked specifically at hallucinations. Some propose an auxiliary loss to encourage unlikely generations from the decoder to receive lower probability \cite{welleck}. However, they only address a specific type of hallucinations, namely repeated n-grams. Other studies have addressed under-translation and over-translation \cite{tucoverage,gargalign}, which are related linguistic phenomena but do not necessarily correspond to hallucinations: under-translation refers to non-translated words, and over-translation refers to words translated multiple times.

Recent work has observed that, when hallucinating, neural MT models tend to over-rely on the generated target tokens and ignore the source sentence \cite{voita-etal-2021-analyzing}. In this paper, we analyse the behaviour of a neural MT model on hallucinated instances {\it wrt} both source and target tokens using a feature attribution method \cite{ig,ib}. Feature attribution methods assign relevance scores to source and target input tokens according to their contribution to the generated output. We observe that a distinctive characteristic of the hallucinated instances is a high entropy of feature attribution scores assigned to the source sentence. Based on this analysis, we devise an auxiliary loss to reduce the amount of hallucinations in MT output. Our experiments show that this loss function successfully mitigates the observed hallucinations.

\begin{table*}
\small
\centering
\begin{tabular}{l|p{12cm}}
\hline
\textbf{Type} & \textbf{Example} \\
\hline
 & {\it Source}: 1936—a făcut apel în justiție contra falsificatorilor operei sale.\\ 
 Oscillatory & Correct Translation: 1936 - filed a lawsuit against the forgers of his work.\\ 
 & {\it Model output}: \textcolor{red}{B6 - B6 - B6 - B6 - B6 - B6 - B6}\\ 
\hline
 & {\it Source}: În iulie 1996, jurnalista Ludmila Gorbul fondează publicația periodică de limba romănă ''Cugetul'' \\
Detached & {\it Correct Translation}: In July 1996, the journalist Ludmila Gorbul founded regular publication on Romanian language.\\
& {\it Model output}: \textcolor{red}{The second subparagraph of Article 6 (1) (c) of Directive 2000 / 29 / EC is to apply.}\\
\hline
 & {\it Source}: Un obicei curent al locuitorilor din Livadia era legat de trenurile personale, care traversau satul și de oprirea lor la halta C.F.R.\\
Mixed & {\it Correct Translation}: A common custom of the inhabitants of Livadia was related to personal trains, which crossed the village and to their stop at the C.F.R.\\
 & {\it Model output}: \textcolor{red}{Primary Copper Smelters - 40 C.F.R. Part 63, Subpart FFFFFFFFFFFFF;}\\
\hline
\end{tabular}
\caption{Examples of hallucination from the MLQE-PE dataset for the Romanian-English language pair. \label{typology_examples}}
\end{table*}

\section{Data}\label{chapter:hallucination_typology}

For our experiments, we use the MLQE-PE dataset \cite{fomicheva2020mlqepe}, which provides manual quality assessments on a continuous $[0..100]$  scale (Direct Assessment -- DA) for $10,000$ sentences per language and access to the neural MT models used to generate the translations (a strong {\tt fairseq} Transformer model). As suggested in previous related work \cite{specia-etal-2021-findings}, this dataset has a high percentage of hallucinated sentences for the Romanian-English (Ro-En) language pair, which was trained on noisier data. We  focus on this language pair in our experiments.

The dataset does not include manual annotation for hallucinations. We use DA scores as a proxy for selecting hallucinated sentences. 
Based on the annotation guidelines, 
we select two subsets: translations with score lower than 25 (lowest quality band) are labelled as containing hallucinations, and translations with scores above 85 (highest quality band) are labelled as not containing hallucinations. 
The resulting dataset contains 567 hallucinated and 3,134 non-hallucinating sentences. 

%
Manual inspection of a sample of sentences with DA scores lower than 25 confirms a clear hallucinating behaviour.
Table~\ref{typology_examples} provides examples of hallucinations of various types, including fully detached sentences that do not have anything in common with the source sentence, oscillatory hallucinations that consist in repeated n-grams \cite{raunak-etal-2021-curious}, as well as mixed cases.



\section{Analysis}\label{chapter:triggers}

First, we look at the input patterns that induce hallucinations. Second, we use feature attribution to gain insight into model behaviour on hallucinated sentences.

\subsection{Hallucination Triggers}

\begin{figure}[tb]
\includegraphics[width=7.5cm]{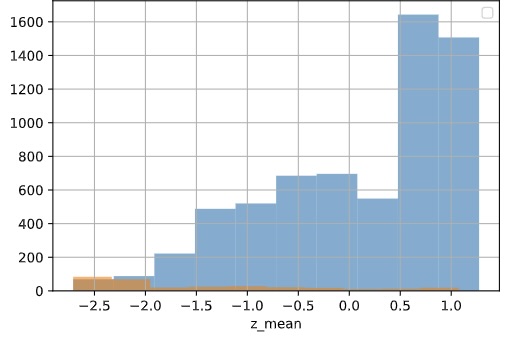}
\caption{Orange: count of instances containing at least one OOV token in the source sentence. Blue: instances containing no OOV tokens in the source sentence. z-mean is the averaged standardised z-score across multiple annotators for a given sentence.}
\label{fig:histogram-contains-UNK}
\end{figure}

Manual inspection of hallucinated sentences in our dataset shows that hallucinations are often caused by the model's lack of robustness given specific inputs. In particular, the model is very sensitive to rare/unknown tokens, such as out-of-vocabulary (OOV) tokens. 
Figure~\ref{fig:histogram-contains-UNK} illustrates this phenomenon: we observe a strong correlation between the presence of at least one OOV token and the hallucination phenomenon, based on the analysis of DA scores. The set of source sentences containing at least one OOV token shows the lowest translation quality scores.

\begin{figure}[tb]
\includegraphics[width=7.5cm]{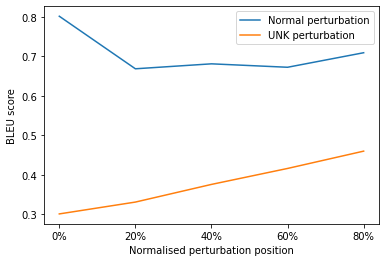}
\caption{Ro-En model. Blue: inserting an in-vocabulary and frequent random token (''e''); orange: inserting the UNK token.}
\label{fig:pertubations-exp-roen}
\end{figure}

\begin{figure}[tb]
\includegraphics[width=7.5cm]{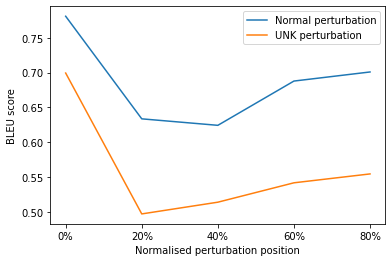}
\caption{Et-En model. Blue: inserting an in-vocabulary and frequent random token (''e''); orange: inserting the UNK token.}
\label{fig:pertubations-exp-eten}
\end{figure}

Having identified this weakness in the model, we wanted to test if by simulating OOV tokens we could induce hallucinations. 
For that, we introduce an out-of-vocabulary token (UNK) in different positions in a subset of 200 sentences with a translation quality score $\geq$ 85, i.e. \textit{clean} sentences (without OOV and correctly translated before). For comparison, we perturb the same set of sentences with a frequent Romanian token "e" in the same positions. The results are shown in Figure \ref{fig:pertubations-exp-roen} for the hallucinating Ro-En model, and in Figure~\ref{fig:pertubations-exp-eten} for a non-hallucinating Estonian-English (Et-En) model. We use BLEU score as a proxy for measuring the degraded quality, in comparison to the reference output (which is the correct translation without any alteration). Perturbations with UNK token have a very strong impact on translation quality. Interestingly, the impact decreases as the token is inserted later in the sentence. In comparison, a perturbation with a commonly used token is less likely to lead to hallucinations. This shows the lack of robustness to rare words in the translation model, which leads to hallucination. In comparison, the Et-En model, under the same OOV perturbation, is much more robust, as the altered BLEU score is lower-bounded by 50 instead of 30 for the Ro-En model.


\subsection{Feature Attribution}\label{chapter:ig_analysis}

Here we study the link between feature attribution scores and the hallucinating behaviour. Feature attribution scores are scalar metrics reflecting the importance of a given input feature for the model prediction. These scores have no ground truth, and are often used as a proxy for model understanding.

We chose Integrated Gradients \cite{ig} as our feature attribution method, as it has been shown effective for other tasks. This method is inspired by the Input$*$Gradients approach, which takes the signed partial derivatives of the output with respect to the input and multiplies them by the input itself. The Integrated Gradients method improves on this by aggregating the gradients along a path from an uninformative baseline input to the actual input: we chose the uninformative baseline to be completely an empty padded source sentence and an empty padded prefix (with only a BOS token), conveying no information. We chose this method as it has been shown to be effective, easy to interpret, and  differentiable, which is a strong requirement as we later devise a training loss function  based on it.

NMT with Transformers is modelled as an auto-regressive task, where each translated token is generated in order. At each generation step, the whole source sentence and all the previously translated tokens are available as inputs to the model. At each time step, we compute attribution scores for each token in the target prefix and in the source. The scores are computed using a sum aggregation over the dimensions of the token embedding.

To investigate whether there are any distinctive patterns in model attributions that would characterise hallucinations, we train an XGBoost classifier to predict if a token (after tokenization and BPE) is hallucinated or not, using a set of features based on the attributions scores: entropy of attributions scores across source tokens, entropy of attributions scores across target tokens and attribution gradients (ignoring padded tokens) relative to the current timestep: at timestep $i$, we analyse the tokens ${t_i, t_{i-1}, ..., t_0}$ and ${s_0, ..., s_{len(s)}}$. Intuitively, we expect an adequate translation to exhibit lower entropy on source attributions with a few source tokens (corresponding to the current target token) having high relevance scores.

Given the lack of datasets with natural hallucinations explicitly annotated at token level, we use our sentence-level dataset (Section \ref{chapter:hallucination_typology}) with a simplifying assumption that all tokens in the hallucinated sentences are labelled as hallucinated. We obtain a validation F1-score of 0.61. Early stopping was used on a distinct validation dataset, with 15 estimators, a max depth of 6, row and columns subsampling ratio of 0.8, minimum child weight of 3.0, lambda and alpha regularization of 1. These parameters were found through randomised grid search, with three values per parameter.
Table ~\ref{tab:halluc_classifier} shows  examples of the predictions of the classifier. 
We observe a strong relationship between the predictions of the classifier and the hallucinating behaviour.
False positives token-level classifications often include tokens ending a word, frequent tokens, and conjunction tokens: for instance, the year \textit{19\textcolor{red}{45}}, frequent tokens such as \textit{\textcolor{red}{at}}, \textit{\textcolor{red}{the}} (in red: wrongly classified as hallucinated). It is probable that these tokens are generated with over-reliance on the target prefixes. 

\begin{table*}
\small
\begin{tabular}{p{15cm}}
\toprule
\textcolor{red}{B ill that the B IP is to be passed, and then the B IP is to be passed , as it is referred to as an} B \textcolor{red}{IP .} \\
It \textcolor{red}{shall} immediately submit \textcolor{red}{, on} simple \textcolor{red}{request by} the \textcolor{red}{Commission , information} on \textcolor{red}{the mesures already taken and planned to comply with this Decision .} \\
N \textcolor{red}{our} re \textcolor{red}{dine is} the main contribut or \textcolor{red}{of the oil} and gas \textcolor{red}{industries} in \textcolor{red}{the DPRK .} \\
The applicant claims that the environmental \textcolor{red}{impact of} the tourist \textcolor{red}{accommodation} is \textcolor{red}{not limited to the overall environmental impact of the tourist accommodation .} \\
\textcolor{red}{In the event of a failure to meet the} requirement \textcolor{red}{of point 1.1. 2.1 ,} the \textcolor{red}{Duc ks concerned have not} been \textcolor{red}{brought forward in the course} of \textcolor{red}{the formal investigation procedure .} \\
B \textcolor{red}{one - off waste ( B B B} C \textcolor{red}{) : half -} scale \textcolor{red}{( B B B C} ) \\
B 6 : B 6 : \textcolor{red}{B 6 : B 6 : B 6 : B 6 : B 6 : B 6 : B 6 :} \\
The second subparagraph is to adopt the basic basic criteria which \textcolor{red}{are to be met by the second} stage \textcolor{red}{.} \\
B \textcolor{red}{6 -} B 6 - \textcolor{red}{B 6 -} B \textcolor{red}{6} - B \textcolor{red}{6} \\
B \textcolor{red}{one} one - off one \textcolor{red}{-} one \textcolor{red}{-} that was \textcolor{red}{then another} - \textcolor{red}{was the B IP .} \\

\midrule
After 19 \textcolor{red}{45} , de colon isation \textcolor{red}{is} taking place and Europe is divided between \textcolor{red}{the} sp heres of the \textcolor{red}{US and the US SR .}\\
In Congres s Hen ry Cl ay was \textcolor{red}{at} the fore front of \textcolor{red}{an} effort to re - authorise \textcolor{red}{.}\\
This massive change has made accumulation and exchange \textcolor{red}{of} experience \textcolor{red}{impossible .}\\
The magnitude of statistical death depends on the number of people affected by the counter \textcolor{red}{measure .}\\
These focused attacks should have destroyed the Ser bian \textcolor{red}{army} at the heart of \textcolor{red}{the} country .\\
An \textcolor{red}{na} and Do sto iv ski are first \textcolor{red}{established} in Berlin , \textcolor{red}{then} moved to D res den \textcolor{red}{.}\\
The population of Malta was estimated at 40 8. \textcolor{red}{000} in July 2011 .\\
III . / J G1 has also lost at least two F w \textcolor{red}{190} for \textcolor{red}{the} same reason .\\
Ben ito Mus sol ini was killed by Italian parti s ans on 28 April .\\
In June 19 44 only 50. 800 t had been produced compared \textcolor{red}{with} 18 \textcolor{red}{0.} 000 t planned .\\
The event at tracted six million \textcolor{red}{vis} itors \textcolor{red}{within} five months .\\
The risks and opportunities for the peaceful use \textcolor{red}{of} nuclear energy have been discussed and \textcolor{red}{have} caused \textcolor{red}{controvers} y\\
German civilian mor als were not a primary \textcolor{red}{objective} for \textcolor{red}{U} SA AF plan ners \textcolor{red}{.}\\
\bottomrule
\end{tabular}{l}
\caption{Hallucinated (top) and non-hallucinated (bottom) translations. The tokens predicted as hallucinations by the token-level binary classifier are marked in red.} 
\label{tab:halluc_classifier}
\end{table*}

\section{Loss Function}\label{chapter:our_loss}

Table~\ref{discrimination_functions_tab_ig} shows the mean values of four features defined in Section \ref{chapter:ig_analysis} on the two subsets of our validation set (correct and hallucinated). 
Gradients is the numerical gradient of the normalised non-zero attributions as a function of the timestep. We then sum the absolute values.  

\begin{table}
\centering
\small
\begin{tabular}{ p{2.5cm} p{2cm} p{2cm} }
\toprule
 & \textbf{Correct} & \textbf{Hallucinated} \\
\toprule
S attr. entropy \raggedright & 3.51\small{$\pm$0.52} \raggedright & 3.91\small{$\pm$0.50} \\  
T attr. entropy \raggedright & 1.73\small{$\pm$0.99} \raggedright & 2.47\small{$\pm$1.15} \\
S att. gradients 
\raggedright & 0.92\small{$\pm$0.22} \raggedright & 1.01\small{$\pm$0.22} \\  
T att. gradients 
\raggedright & 0.74\small{$\pm$0.33} \raggedright & 0.74\small{$\pm$0.32} \\  
\bottomrule
\end{tabular}
\caption{Entropy and gradients of the attribution scores for Source (S) and Target (T) for the hallucinated and correct translations in our dataset. Mean and standard deviation are computed for each portion of the dataset (completely correct and completely hallucinated).} 
\label{discrimination_functions_tab_ig}
\end{table}

Overall, the mean values are higher for hallucinated tokens, except for attribution gradients on the target prefix. The most discriminative feature seems to be the entropy of attribution values on the source tokens. 
Based on this observation, we propose the following learning objective for MT training:
\begin{equation}\label{eq:loss}
    \mathcal{L} = \mathcal{L}_{CE} + \lambda \mathcal{L}_{Attr}
\end{equation}
where $\mathcal{L}_{CE}$ is the standard cross-entropy (CE) loss with label smoothing, and $\mathcal{L}_{Attr}$ is the proposed attribution-based loss. We define $\mathcal{L}_{Attr}$ as the entropy of the attribution values on the source tokens averaged across generation time steps:
\begin{equation} \label{eq:custom_loss_def}
\mathcal{L}_{Attr} = \mathcal{H}[IG^*(X_{source})]
\end{equation}
where $IG^*$ corresponds to the Integrated Gradients attribution method. This method has a prohibitive computational complexity to be used at training time for a generation task. To reduce computational complexity, we use an approximation, with only one step in the comparison between the input and the baseline representation: 
\begin{equation} \label{eq:simple_gradients_def}
    IG^*_i(x) := (x_i -x_i^{'}) \frac{\partial F(x)}{\partial x_i}
\end{equation}
with $x_i$ being the input, $x_i'$ the uninformative baseline input (empty source sentence and BOS target sentence) and $F$ the model function.
$x_i^{'}$ is the vector generated by the following sequences of tokens: $T_{source} = ([PAD], ... [PAD])$ and $T_{prefix} = ([BOS], [PAD], ..., [PAD])$.

Teaching the model to decrease the entropy of the source attributions is justified by the statistics shown in Table~\ref{discrimination_functions_tab_ig}, as this value is lower for non-hallucinated sentences. Moreover, it also encourages the model the narrow the focus on tokens, hence making the influence of perturbation tokens less important along the entire sentence, as in the perturbations experiments in Section~\ref{chapter:triggers}.

To test whether our method reduces the amount of hallucinated sentences, we fine-tune the base Ro-En translation model using the proposed auxiliary loss. For comparison we keep training the base model with the standard CE loss under the same process: same hyper-parameters and early stopping. We used Adam as optimiser, the learning rate was set to continue from the last epoch before fine-tuning, the scheduler was reset (inverse square root), weight decay was set to 10e-4, label smoothing weight to 0.1 and dropout ratio to 0.3. The maximum number of epochs was set to 30. The model was trained with the {\tt fairseq} library, same as the base model. We set the weight of our loss to 5 (versus 1 for the cross-entropy): this was to keep the same order of magnitude between the two values during training.

We evaluate the performance of the models on the full Ro-En MLQE-PE dataset, as well as specifically on the hallucinated and non-hallucinated subsets. Since the dataset does not provide reference translations, we back-translate the generated English MT outputs into Romanian using Google Translate API and compute BLEU score between the back-translated Romanian sentences and the original source sentences from the dataset.

\begin{table}
\centering
\small
\begin{tabular}{ p{1.5cm} r r r }
\toprule  
\textbf{Model} & \textbf{Overall} & \textbf{Halluc.} & \textbf{Non-Halluc.} \\
\toprule
\textbf{Baseline} \raggedright & 28.90 & 10.71 & 35.29 \\
\textbf{Fine-tuned} \raggedright & 38.25 & 24.46 & 44.90 \\
\bottomrule
\end{tabular}
\caption{BLEU scores of the model fine-tuned with the proposed auxiliary loss compared to the base model on different subsets of data. } 
\label{sacrebleus_finetuning}
\end{table}

Table~\ref{sacrebleus_finetuning} shows the results. We observe a very strong improvement of almost 10 BLEU points from fine-tuning with the proposed loss on the entire test set (Overall - 7,000 sentences). The improvement is even more evident for the set of source sentences with the lowest initial translation quality, which we hypothesised was due to a high incidence of hallucinations (Halluc.). The fine-tuned model with our additional loss successfully mitigates the hallucination behaviour, compared to the baseline model fine-tuned with the original loss under the same conditions. 

\begin{table}
\centering
\small
\begin{tabular}{l p{4.4cm}}
\toprule
Source & În iulie 1996, jurnalista Ludmila Gorbul fondează publicația periodică de limba romănă ''Cugetul'' \\
Base model & \textcolor{red}{(RO) Mr President, in July 1996, the official} Ludmila Gorul founded the regular publication of the Romanian language, \textcolor{red}{which was published on the Official Journal of the European Union}. \\
Our model & In July 1996, the journalist Ludmila Gorbul founded regular publication on Romanian language.\\
\bottomrule
\end{tabular}
\caption{Example of hallucinated MT output by the base model and the improved version by our model.}
\label{tab:example}
\end{table}

\begin{figure*}
\centering
\includegraphics[width=16cm]{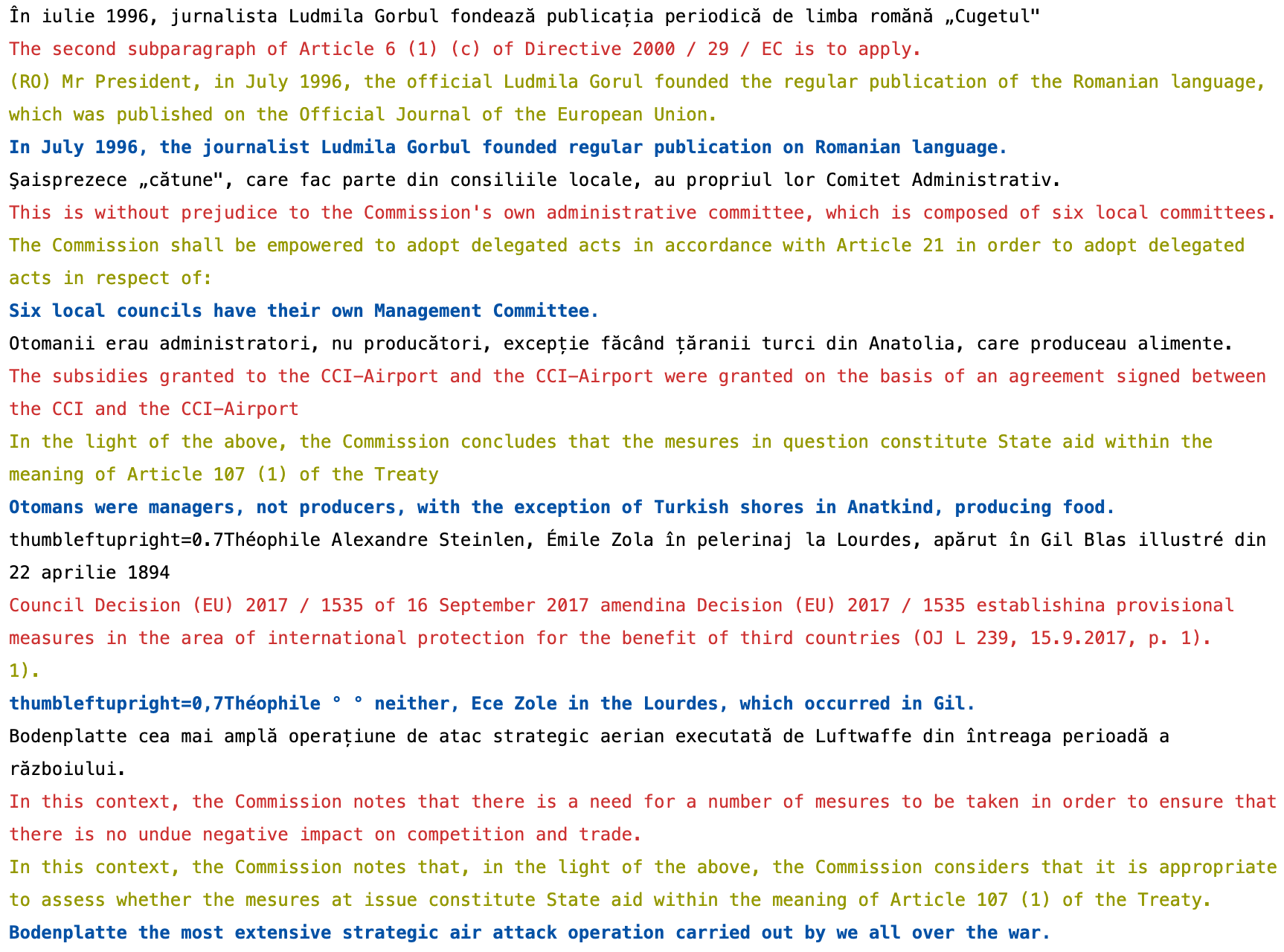}
\caption{The fine-tuned model consistently outputs less hallucinations. In black: the source sentence. In red: the model output before fine-tuning (Transformer). In green: the baseline (cross-entropy with label-smoothed cross-entropy). In blue: our fine-tuned model.}
\label{fig:hallucinations_corrected}
\end{figure*}

\begin{figure*}
\centering
\includegraphics[width=16cm]{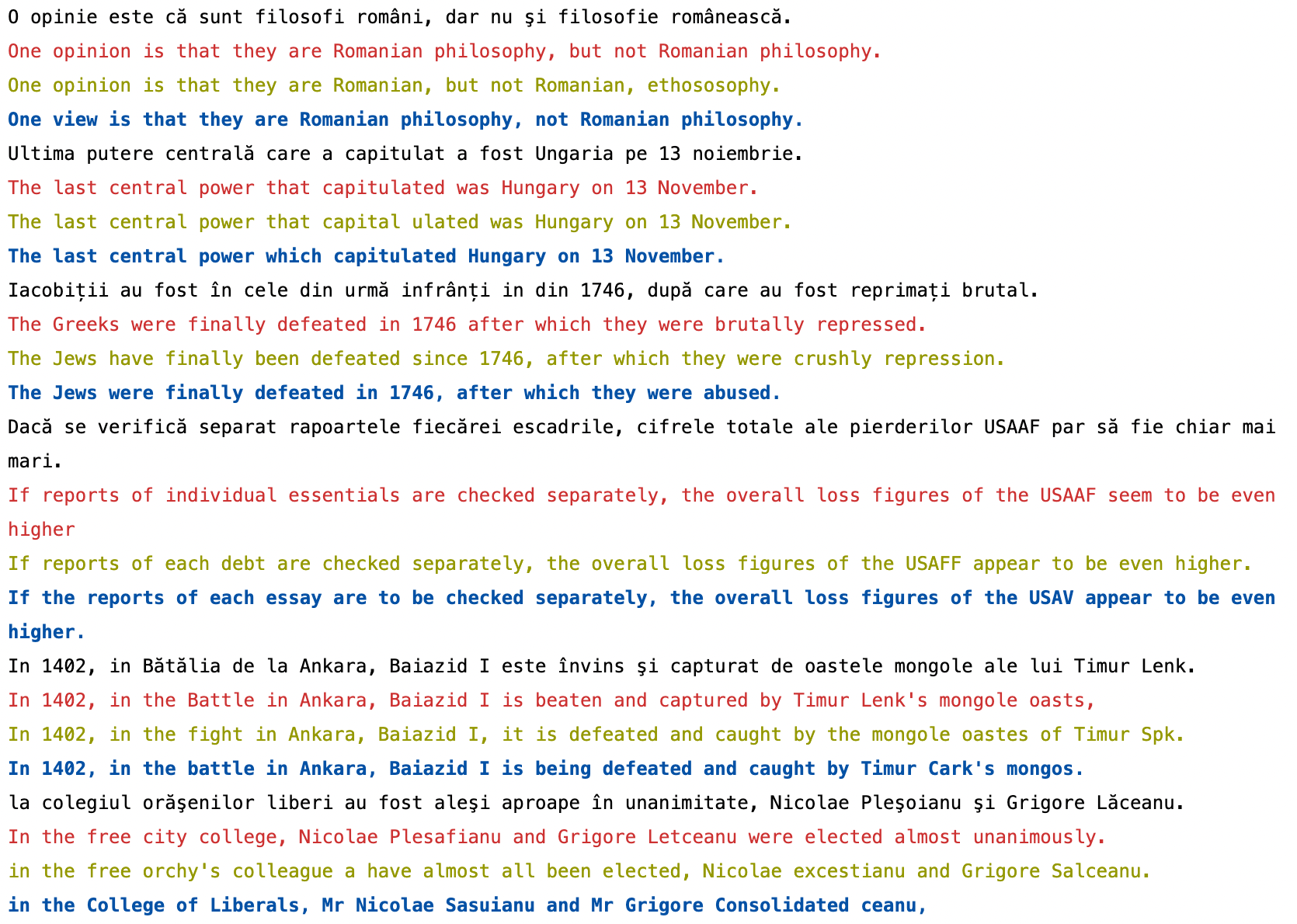}

\caption{The fine-tuned model consistently conveys the same translations on previously not hallucinated sentences. In black: the source sentence. In red: the model output before fine-tuning (Transformer). In green: the baseline (cross-entropy with label-smoothed cross-entropy). In blue: our fine-tuned model.}
\label{fig:good_sentences}
\end{figure*}

Table \ref{tab:example} shows an example of improvement on a previously hallucinated translation by the base model. Figure~\ref{fig:hallucinations_corrected} is a comparison of the output of different models for various source sentences (in black): the initial translation from the MLQE dataset (in red), the translation from the baseline model (fine-tuned with the same setup as our method) (in orange), and the translation from the model fine-tuned with the proposed auxiliary loss function (in blue). Figure~\ref{fig:good_sentences} illustrates the same comparison on previously perfectly translated sentences. In this case, all the outputs should be close, showing that our proposed approach is not harming outputs that would have been correct without the additional loss function.
In particular, despite the presence of an OOV token (the first quotation mark in the source sentence), our model successfully recovers the main meaning of the source sentence in its translation, even if the coverage is partial (the word ''Cugetul'' is missing). 

\section{Conclusions}

Our work studies the hallucinations phenomenon on a Romanian-to-English translation model, using a standard Transformer architecture. First, we analysed the observed hallucinations in order to understand potential triggers that let to them.
We showed that specific patterns from the source sentence could induce hallucinations, which is an interesting take on the hallucination problem from a data/model robustness perspective. 
Then, we studied the behaviour of the hallucinating model using feature attribution methods to investigate whether this behaviour can be quantified in a differentiable way that could be used to inform the design of new loss functions for better model training.  
We found that the entropy of the attribution values on the source tokens is effective at discriminating hallucinating from non-hallucinating tokens.  
Based on these findings, we proposed a novel and intuitive loss function that successfully increases the robustness of the model to the observed trigger patterns and hence decreases hallucinations occurrences.

In future work, we plan to study whether our approach generalises to other language pairs and hallucinations caused by other triggers. 

\bibliography{eamt23,anthology,custom}
\bibliographystyle{eamt23}

\end{document}